\documentclass[10pt,twocolumn]{article}

\usepackage[a4paper,margin=1.8cm]{geometry}
\usepackage[T1]{fontenc}
\usepackage[utf8]{inputenc}
\usepackage{newtxtext,newtxmath}   
\usepackage{microtype}
\usepackage{abstract}

\usepackage{cite}
\usepackage{amsmath,amsfonts}
\usepackage{graphicx}
\usepackage{algorithm}
\usepackage{algpseudocode}
\usepackage{textcomp}
\usepackage{caption}
\usepackage{float}
\usepackage{booktabs}
\usepackage{diagbox}
\usepackage{multirow}
\usepackage{gensymb}
\usepackage{dsfont}
\usepackage[table,xcdraw]{xcolor}
\usepackage{makecell}
\usepackage{adjustbox}
\usepackage[hidelinks]{hyperref}

\definecolor{darkgreen}{RGB}{0,100,0}
\newcommand{\inc}{\textcolor{red}{$\uparrow$}}
\newcommand{\dec}{\textcolor{darkgreen}{$\downarrow$}}
\DeclareUnicodeCharacter{FF0C}{,}

\title{\textbf{Perturb-and-Restore: Simulation-driven Structural Augmentation Framework for Imbalance Chromosomal Anomaly Detection}}

\author{
Yilan Zhang$^{1,2,*}$, Hanbiao Chen$^{3,*}$, Changchun Yang$^{1,2,*}$, Yuetan Chu$^{1,2}$,\\
Siyuan Chen$^{1,2}$, Jing Wu$^{3}$, Jingdong Hu$^{4}$, Na Li$^{4}$, Junkai Su$^{4}$,\\
Yuxuan Chen$^{4}$, Ao Xu$^{4}$, Xin Gao$^{1,2,\dagger}$, Aihua Yin$^{3,\dagger}$\\[0.5em]
\small $^1$ Computer Science Program, Computer, Electrical and Mathematical Sciences and Engineering Division,\\
\small King Abdullah University of Science and Technology (KAUST), Thuwal 23955-6900, Saudi Arabia\\
\small $^2$ Center of Excellence on Smart Health (KCSH) and Center of Excellence for Generative AI,\\
\small KAUST, Thuwal 23955-6900, Saudi Arabia\\
\small $^3$ Guangdong Provincial Maternal and Child Health Hospital, Guangzhou 511442, China\\
\small $^4$ Smiltec, Suzhou 215125, China\\[0.3em]
\small $^*$ Equal contribution\\
\small $^\dagger$ Correspondence: xin.gao@kaust.edu.sa; yinaihua@vip.126.com
}

\date{}

\begin{document}
\twocolumn[
\maketitle
\begin{onecolabstract}
Detecting structural chromosomal abnormalities is crucial for accurate diagnosis and management of genetic disorders. However, collecting sufficient structural abnormality data is extremely challenging and costly in clinical practice, and not all abnormal types can be readily collected. As a result, deep learning approaches face significant performance degradation due to the severe imbalance and scarcity of abnormal chromosome data. To address this challenge, we propose a \textbf{Perturb-and-Restore (P\&R)}, a simulation-driven structural augmentation framework that effectively alleviates data imbalance in chromosome anomaly detection. The P\&R framework comprises two key components: (1) \textbf{Structure Perturbation and Restoration Simulation}, which generates synthetic abnormal chromosomes by perturbing chromosomal banding patterns of normal chromosomes followed by a restoration diffusion network that reconstructs continuous chromosome content and edges, thus eliminating reliance on rare abnormal samples; and (2) \textbf{Energy-guided Adaptive Sampling}, an energy score-based online selection strategy that dynamically prioritizes high-quality synthetic samples by referencing the energy distribution of real samples. To evaluate our method, we construct a comprehensive structural anomaly dataset consisting of over \textbf{260,000} chromosome images, including \textbf{4,242} abnormal samples spanning \textbf{24 categories}. Experimental results demonstrate that the P\&R framework achieves state-of-the-art (SOTA) performance, surpassing existing methods with an average improvement of \textbf{8.92\%} in sensitivity, \textbf{8.89\%} in precision, and \textbf{13.79\%} in F1-score across all categories.
\end{onecolabstract}
\vspace{0.8em}
]

\section{Introduction}
Chromosomes are nuclear structures composed of DNA and proteins that carry the genetic material essential for human inheritance and development~\cite{sharma2014chromosome}. Human somatic cells typically contain 46 chromosomes arranged in 23 pairs (22 autosomes and XY/XX sex chromosomes). Karyotyping—the systematic arrangement and analysis of chromosomes—is a cornerstone technique in cytogenetics, widely used to detect chromosomal abnormalities in cancers, inherited disorders, and other conditions~\cite{sharma2014chromosome, mitelman2010mitelman}. For instance, the Philadelphia chromosome in chronic myeloid leukemia serves as a key diagnostic and therapeutic biomarker~\cite{tefferi2015myeloproliferative}.

The chromosomal abnormalities are broadly classified into numerical and structural changes. Specifically, numerical abnormalities arise from alterations in chromosome number, such as the presence of an extra chromosome (trisomy) or the absence of a chromosome (monosomy), which can result in genetic diseases, for example, Down syndrome, Edward syndrome and Turner syndrome~\cite{jaikrishan1999genetic, winther2004chromosomal}， and can be detected by chromosome counting~\cite{kang2024chromosome}. In contrast, structural abnormalities involve more complex alterations—such as deletions, duplications, translocations, and inversions—which affect the internal architecture of chromosomes. Unlike numerical changes, structural abnormalities are often subtle, highly variable, and more difficult to detect, particularly when affecting small genomic regions. Their identification remains a major challenge for standard cytogenetic techniques due to their stochastic presentation and limited visual distinguishability~\cite{alkan2011genome, carvalho2016mechanisms}.

Traditional karyotyping is a labor-intensive and time-consuming process that requires significant expertise, making it prone to human error. To address these limitations, computer-aided automated detection has been increasingly incorporated into karyotyping workflows, enabling more rapid and accurate diagnoses in clinical settings. Deep learning (DL) pipelines have achieved notable advances in various chromosome analysis tasks, including segmentation ~\cite{karvelis2006watershed, mei2020adversarial, liu2024algorithm}, classification ~\cite{sharma2017crowdsourcing,jindal2017siamese,abid2018survey, huang2025notac}, straightening ~\cite{zheng2022chrsnet,li2023masked}, and trisomy detection ~\cite{al2020automated}. However, research on the detection of structural chromosomal abnormalities remains limited. Two major challenges hinder progress in this area. 

First, structural chromosomal abnormalities are inherently rare and often extremely challenging and costly to collect in sufficient quantities. As a result, datasets exhibit severely imbalanced, long-tailed distributions within each chromosome class—abnormal instances are significantly fewer compared to abundant normal ones. This imbalance significantly degrades the performance of deep learning models, which tend to overfit to the dominant class~\cite{tao2023local}. Furthermore, existing methods generally focus on specific types of abnormalities and often overlook the diversity and imbalance among different categories of abnormalities ~\cite{cox2022automated, bechar2023highly}. Consequently, these approaches exhibit limited generalizability, struggling to effectively model and detect the broad spectrum of structural chromosomal abnormalities encountered in realistic clinical diagnostic scenarios.

Second, data synthesis-based augmentation offers a direct solution to data imbalance by generating rare anomalies. While generative models can in principle produce abundant abnormal samples, most existing synthesis techniques are task-specific—e.g., 3D scenes~\cite{gaidon2016virtual}, weather changes~\cite{guo2022domain}, or MR images~\cite{zhou2020hi, kaur2021mr}—and are not well-suited to capturing the fine-grained structural variability of chromosomes. The scarcity of real abnormal chromosome data further hinders the training of generative models, which struggle to learn the underlying distribution of rare and diverse anomaly types~\cite{yi2019generative}. Moreover, the absence of ground-truth annotations for synthetic samples introduces an additional bottleneck: without reliable evaluation metrics, it becomes difficult to judge the authenticity or utility of generated data. Therefore, two key challenges remain unresolved: how to generate structurally realistic and diverse synthetic anomalies in the absence of sufficient real abnormal data, and how to implicitly assess and dynamically prioritize high-quality synthetic samples during training to maximize their utility for downstream anomaly detection.

\begin{figure*}
    \centering
    \includegraphics[width=1\linewidth]{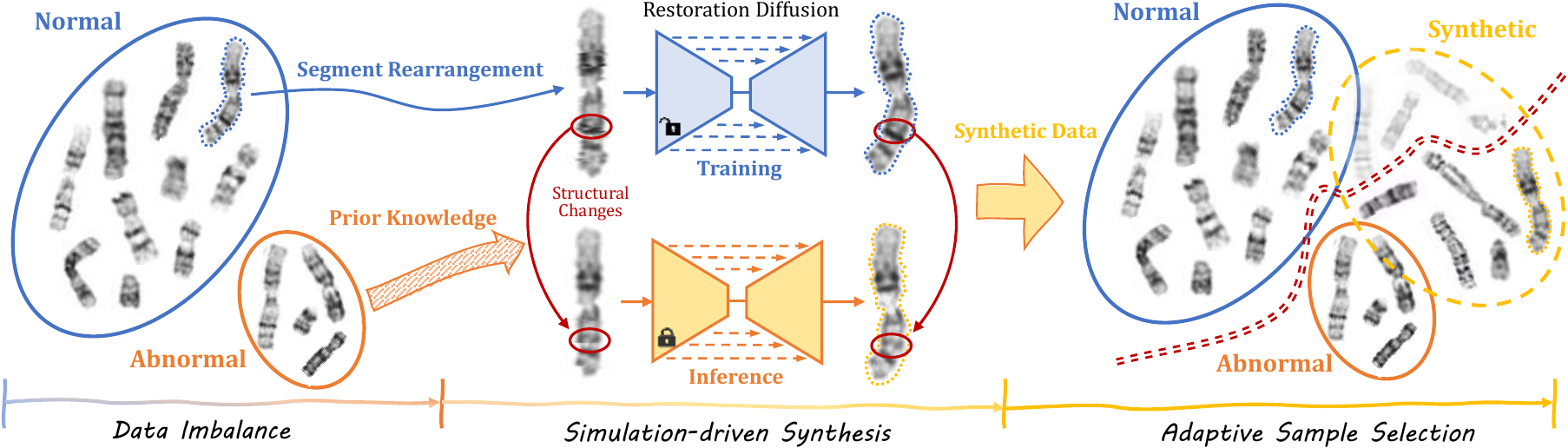}
    \caption{Thoughts of Perturb-and-Restore (P\&R). P\&R mitigates the impact of long-tailed chromosomal distributions by simulating diverse structural anomalies from normal chromosomes. Guided by prior knowledge, structural perturbations are introduced to specific regions and refined via diffusion-based restoration. An energy-guided sampling strategy then selects high-quality synthetic anomalies to enhance the training of robust anomaly detectors.}
    \label{fig:introduction}
\end{figure*}

To address the above issues, we propose \textbf{Perturb-and-Restore (P\&R)}, a novel simulation-driven structural augmentation framework designed to tackle data imbalance in structural chromosomal anomaly detection. The P\&R framework consists of two main components: (1) a \textbf{S}tructural \textbf{P}erturbation and \textbf{R}estoration \textbf{S}imulation (\textbf{SPRS}), and (2) \textbf{E}nergy-guided \textbf{A}daptive \textbf{S}ampling (\textbf{EAS}). Firstly, the \textbf{SPRS} addresses the limited availability of abnormal chromosome data by generating synthetic anomalies through an innovative perturbation-restoration approach.  Specifically, chromosomal banding patterns of normal chromosomes are first perturbed by rearranging chromosome segments and introducing structural anomalies such as deletions, duplications, inversions, and replacements, guided by prior genetic knowledge. However, these perturbations result in discontinuities and misalignments in chromosome structures. To overcome this issue, we introduce a restoration diffusion network trained via a score-based diffusion model ~\cite{song2020score} on pairs of original and rearranged normal chromosome images, effectively reconstructing continuous and realistic chromosome content and edges. This indirect generation method leverages abundant normal chromosome data, ensuring high-quality synthetic anomalies without reliance on scarce abnormal samples.

Secondly, the \textbf{EAS} component employs energy scores to dynamically prioritize synthetic samples during training. By explicitly creating an energy gap—assigning lower energy scores to normal chromosomes and higher scores to abnormal ones—the strategy dynamically selects synthetic samples whose energy scores are close to those of real anomalies. Given that standard metrics cannot directly measure synthetic abnormal data quality, this adaptive sampling strategy iteratively exposes the detection model to newly sampled, high-quality synthetic abnormal samples, enhancing the network’s discriminative capability and aligning the synthetic data distribution closely with real anomalies, thus significantly improving anomaly detection performance. Notably, our approach is broadly applicable to diverse structural chromosomal abnormalities and not confined to specific types. P\&R offers a generalizable paradigm for tackling data imbalance in rare genetic disorder diagnosis. 

\section{Related Works}
\subsection{Automated Chromosome Classification}
Automated chromosome classification has progressed from traditional machine learning methods (Bayes classifier and support vector machines (SVMs))~\cite{gilbert1966computer,carothers1994computer,markou2012automatic,mashadi2007direct} to modern deep learning models that achieve near-expert performance on the 24-class task~\cite{qin2019varifocal}. Advanced architectures, including siamese network ~\cite{jindal2017siamese} and hybrid models combining CNNs with recurrent neural networks (RNNs) ~\cite{sharma2018automatic} or self-attention mechanisms ~\cite{xia2023karyonet}, have further improved accuracy and robustness. More recently, research has expanded to detecting abnormal chromosomes, including trisomy identification ~\cite{al2020automated}, dicentric chromosome detection ~\cite{shen2019dicentric}, and classification of specific structural anomalies such as deletions (e.g., del(5) ~\cite{bechar2023automatic}) and translocation (e.g.t(9;22)~\cite{chen2024br}). Cox et al. ~\cite{cox2022automated} applied cyclical learning rates to train VGG ~\cite{simonyan2014very} and ResNet ~\cite{he2016identity} on 12 recurrent abnormal chromosome classes, while ChroSiameseAD ~\cite{bechar2023highly} employed a siamese network with contrastive learning to detect deletion del(5) inversion inv(3). Although these studies demonstrate the growing focus on structural anomaly detection, they are limited to specific types of abnormalities and do not address the long-tailed distribution issue caused by the scarce abnormal data in real-world scenarios. To overcome this limitation, we construct a large-scale dataset spanning all 24 chromosome categories and propose the Perturb-and-Restore (P\&R) framework. P\&R simulates abnormal chromosomes and employs energy-guided sampling to prioritize high-quality synthetic data during training, enabling robust detection under long-tailed imbalance.

\subsection{Long-tailed Learning}
The long-tailed problem—where a few dominant classes have abundant samples while others are scarce—poses a major challenge to training robust deep models~\cite{kang2020exploring}. Existing methods can be categorized into re-balancing strategies, module improvement, and data augmentation ~\cite{zhang2023deep, zhang2023ecl}. Re-balancing methods include re-sampling techniques, such as over-sampling minority classes ~\cite{ando2017deep} or under-sampling majority classes ~\cite{he2009learning}, and re-weighting losses like focal loss ~\cite{ross2017focal} and LDAM-DRW loss ~\cite{cao2019learning}, which adjust class contributions during training. Module improvement methods focus on enhancing feature representations and classifiers. For instance，decoupled training ~\cite{kang2019decoupling,zhou2020bbn} improves performance by using different network branches to optimize feature extraction and classification. And self-supervised methods, such as balanced contrastive learning (BCL) ~\cite{zhu2022balanced} and targeted supervised contrastive learning (TSC) ~\cite{li2022targeted}, learn robust representations by aligning positive sample pairs and repelling negative ones. Additionally, data augmentation techniques ~\cite{li2021metasaug, khan2024review,chen2025zero}, including synthetic data generation, address class imbalance by creating additional samples for minority classes. However, most are domain-agnostic and overlook the structural complexity of chromosomal anomalies. In this work, we propose a tailored simulation-based augmentation strategy that leverages domain-specific prior knowledge to generate diverse and realistic anomalies without relying solely on scarce abnormal data.

\section{Methods}

We propose the \textbf{Perturb-and-Restore (P\&R)} framework for structural chromosomal anomaly detection, comprising two key components: (1) \textbf{Structure Perturbation and Restoration Simulation (SPRS)} for generating realistic abnormal chromosomes, and (2) \textbf{Energy-guided Adaptive Sampling (EAS)} for selecting high-quality synthetic samples. Given a dataset \( D = \cup_{c \in C} D_c \), where \( C = \{1, \dots, 22, X, Y\} \) denotes the 24 chromosome categories and \( D_c = \{D_c^n, D_c^{ab}\} \) includes normal and abnormal samples, SPRS simulates abnormalities by rearranging segments of normal chromosomes. A score-based diffusion model is then used to restore structural continuity, producing high-quality synthetic anomalies \( \hat{D}_c^{ab} \). Notably, the model is trained in an abnormal label-free manner, thus it can leverage both normal and abnormal chromosomes. In EAS, we propose an online sampling strategy to further enhance performance by iteratively selecting synthetic samples from \( \hat{D}_c^{ab} \) whose energy scores align with those of real anomalies. By referencing the energy score distribution of real anomalies, EAS prioritizes synthetic samples with similar characteristics, thereby improving anomaly detection performance. The details will be described in the following sections. Details are described in the following sections.

\begin{figure*}
    \centering
    \includegraphics[width=1\linewidth]{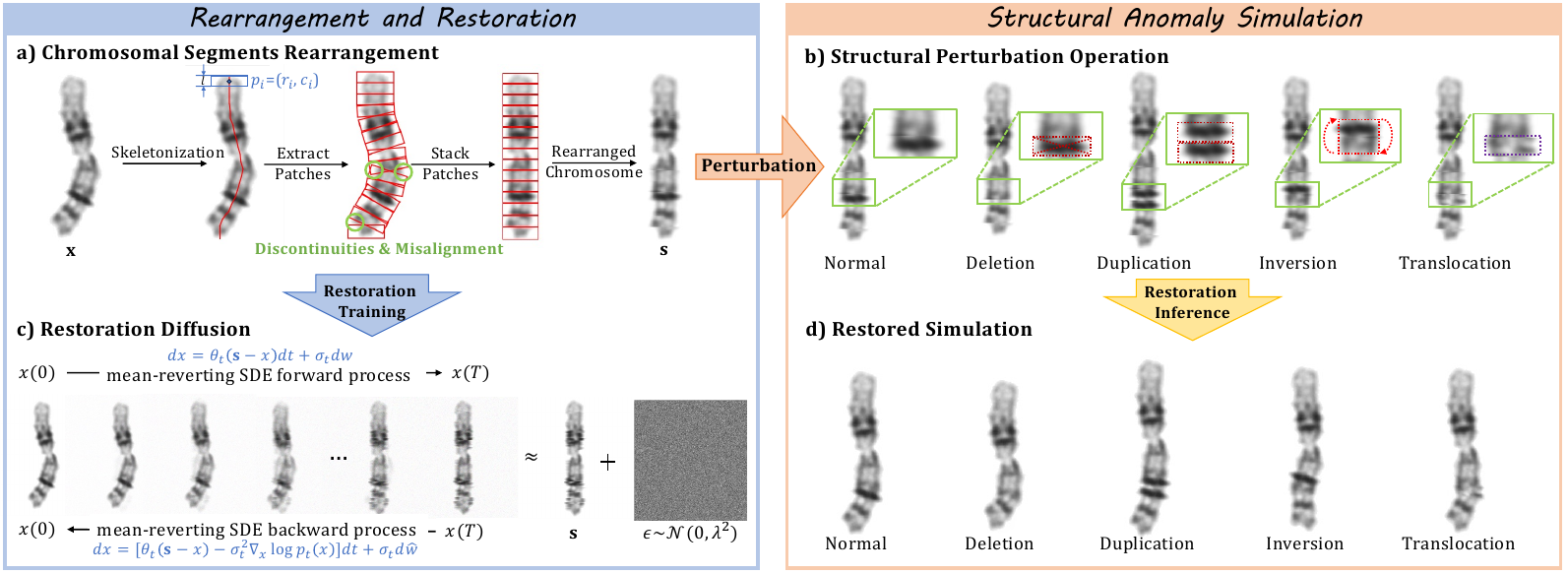}
    \caption{The flowchart of the Structure Perturbation and Restoration Simulation (SPRS) module: (a) Chromosome skeletons are extracted and segmented into rectangular patches. (b) Structural perturbation operations are applied to rearranged chromosomes using prior knowledge.(c) The rearranged chromosomes and their original counterparts are paired to train a diffusion-based restoration model that treats rearrangement as image degradation. (d) The trained restoration model refines these synthetic anomalies by restoring realistic edges and content, generating highly realistic abnormal samples.}
    \label{fig:rearrangement}
\end{figure*}

\subsection{Structure Perturbation and Restoration Simulation (SPRS)}\label{sec:sprs}
\subsubsection{Chromosomal Segments Rearrangement}
Rearrangement serves as a preliminary step for both structural abnormal simulation and training data construction for the restoration model. The procedure is illustrated in Fig. \ref{fig:rearrangement}. Given a chromosome image $\mathbf{x}$, we first extract the chromosome mask using binary thresholding and apply the medial axis thinning algorithm ~\cite{lee1994building} to obtain its skeleton. To extract a clean medial axis, we merge multiple paths in the skeleton that share common nodes or are in close proximity and identify the longest path as the final axis. Based on this axis, the chromosomes can be represented as a series of rotated rectangular segments centered along the medial axis. Specifically, we sample the axis at interval $l$, obtaining points $\mathbf{p}=\{p_1,p_2,\dots,p_n\}$ where $p_i=(r_i,c_i)$ denotes the center coordinates of the $i$-th patch and $||p_{i+1}-p_{i}||=l$. By stacking these patches without overlap, we construct a rearranged chromosome $\mathbf{s} = \{s_1,s_2,\dots,s_n\}$ where $s_i$ is centered at $p_i$. By rearranging these patches, structural perturbations can be simulated to mimic various types of chromosomal abnormalities.

\subsubsection{Structural Perturbation Operations}\label{sec:structural_perturbation}
 
To simulate diverse chromosomal abnormalities, we apply structural perturbation operations to the stacked rectangular patches $\mathbf{s}$, mimicking common variants such as deletions, duplications, inversions, and translocations:

\begin{itemize}
\item Deletion: A patch \(s_j\) is removed:
\begin{equation}
    \mathbf{s'} = \mathbf{s} \setminus \{s_j\},
\end{equation}

\item Duplication: A patch \(s_j\) is duplicated and reinserted consecutively:
\begin{equation}
\mathbf{s'} =\{s_1, \dots, s_j, s_j, \dots, s_n\},
\end{equation}

\item Inversion: A subset of patches \(\{s_j, s_{j+1}, \dots, s_m\}\) is reversed:
\begin{equation}
\mathbf{s'} = \{s_1, \dots, s_m^\text{inv}, \dots, s_j^\text{inv}, \dots, s_n\},
\end{equation}
where \(s_j^\text{inv}\) denotes the inverted patch.

\item Translocation:
a segment 
$s_{j}$ from a chromosome of class 
$c$ is replaced with a segment $s'_{j}$ from a chromosome belonging to a different class $c'$:
\begin{equation}
\mathbf{s'} = \{s_1, \dots, s_j \mapsto s'_j, \dots, s_n\},
\end{equation}
\end{itemize}

In practice, highly curved chromosomes may lead to information loss or patch redundancy after stacking, reducing perturbation quality. To mitigate this, we propose the medial axis cosine (MAC) score, which quantifies global curvature by measuring the alignment between local and global skeleton directions. Given $M$ sampled points along the medial axis, MAC is computed as:
\begin{equation}
\begin{aligned}
\mathbf{d}_{\text{g}} &= \frac{(r_{b} - r_{t}, c_{b} - c_{t})}{\|\mathbf{d}_{\text{g}}\|}, \\
\mathbf{d}_{i} &= \frac{(r_{i} - r_{i-1}, c_{i} - c_{i-1})}{\|\mathbf{d}_{i}\|}, \\
\delta_{i} &= 1 - \mathbf{d}_{i} \cdot \mathbf{d}_{\text{g}}, \\
\text{MAC} &= \left( 1 - \frac{1}{M} \sum_{i=2}^{M} |\delta_{i}| \right) \times 100
\end{aligned}
\end{equation}
 where, $(r_t,c_t)$ and $(r_b,c_b)$ denote the topmost and bottommost skeleton points. A higher MAC score (closer to 100) indicates a straighter chromosome. We retain only chromosomes with MAC $>$ 85 for simulation, ensuring structure quality. However, due to natural curvature and rearrangement, the outputs may still contain visual artifacts such as edge discontinuities or internal misalignment—motivating the need for further refinement via image restoration, addressed in the next section.

\subsubsection{Abnormal Label-free Rearrangement Restoration}
 To address the above issue and enhance realism, we formulate the task as an image restoration problem. Specifically, we treat the rearranged chromosome $\mathbf{s}$ as a degraded version of the original $\mathbf{x}$, where the reassembly along the medial axis introduces geometric distortions and partial information loss. Recent advances in generative modeling, particularly diffusion models, have demonstrated strong capabilities in high-fidelity image synthesis and restoration. Instead of employing vanilla diffusion, we use a variant of stochastic differential equation (SDE)-based diffusion model, mean-reverting SDE~\cite{luo2023image}, as it does not depend on the knowledge of task-specific restoration and is well-suited to our objective. Given a pair of chromosome images--$\mathbf{x}$ and its rearranged counterpart $\mathbf{s}$, the forward process diffuses $\mathbf{x}$ into $\mathbf{s}$ with added Gaussian noise, while the reverse-time SDE restores the natural curvatures and edges.
 
Let $\{x(t)\}_{t=0}^T$ represent the continuous diffusion trajectory over time $t \in [0,T]$. The initial condition $x(0)$ is set as $\mathbf{x}$, while the terminal state $x(T)$ follows a Gaussian distribution in its noisy form, given by $\mathbf{s} +\epsilon$, where $\epsilon \sim \mathcal{N}(0, \lambda^2)$ denotes the fixed noise. The \textbf{forward process} of SDE is formulated as:
\begin{equation}\label{eq:forward}
     dx = \theta_t(\mathbf{s}-x)dt +\sigma_tdw,
 \end{equation}
 where $w$ is a standard Wiener process, $\theta_t$ is a time-dependent drift coefficient controlling the mean reversion speed, and $\sigma_t$ is the stochastic volatility parameter governing noise injection. The term $\theta_t(\mathbf{s}-x)dt$ represents the drift, which pulls $x$ to $\mathbf{s}$, while the term $\sigma_tdw$ represents the diffusion, introducing randomness into the process. We set the coefficients to satisfy $2\lambda^2=\sigma_t^2/\theta_t$, allowing a closed-form solution of Eq.\eqref{eq:forward}. 
 
To restore the rearranged chromosomes to their original form, the \textbf{reverse-time representation} ~\cite{song2020score} is formulated as:
\begin{equation}
    dx = [\theta_t(\mathbf{s}-x)-\sigma_t^2 \nabla_x\log p_t(x)]dt + \sigma_t d\hat{w},
\end{equation}
where $\nabla_x\log p_t(x)$ is the score function. During training, since the ground-truth $x(0)$ is accessible, the score function at each time step $t$ can be computed. Using the parameterization trick, we sample $x(t)=m_t+\sqrt{v_t}\epsilon_t$ from standard Gaussian noise $\epsilon_t \sim \mathcal{N}(0,I)$, where $m_t$ and $v_t$ are the mean and variance of the marginal distribution $p_t(x)$ at any time $t$, which are calculated during the forward process. The score in terms of the noise is then computed as:
\begin{equation}
    \nabla_x\log p_t(x)=-\frac{\epsilon_t}{\sqrt{v_t}}
\end{equation}

To approximate the noise $\epsilon_t$, we employ a U-Net based latent diffusion model, denoted as  $\tilde{\epsilon}_\phi(x(t),\mathbf{s},t)$, trained using the objective function from DDPM ~\cite{ho2020denoising}:
\begin{equation}
    \mathcal{L}_{\gamma}(\phi):=\sum_{i=1}^{T}\gamma_{i}\mathbb{E}[||\tilde{\epsilon}_\phi(x_i,\mathbf{s},i)-\epsilon_i||]
\end{equation}
where $\gamma_i$ is the positive weight at the discrete time point $i$

For generating synthetic structurally abnormal chromosomes, we start from a simulated chromosome $\mathbf{s'}$ and update $x(t)$ using the reverse SDE. At each time step $t$, the trained noise network $\tilde{\epsilon}_\phi(x(t),\mathbf{s'},t)$ estimates the noise and progressively denoises the sampled chromosome image. This process effectively corrects edge discontinuities and content misalignments, ensuring the restoration of high-quality chromosomal structures. Ultimately, we obtain the synthetic dataset $\hat{D}_c^{ab}$, which enhances data realism and serves as a robust augmentation resource for training anomaly detection models.

\begin{figure*}
    \centering
\includegraphics[width=1\linewidth]{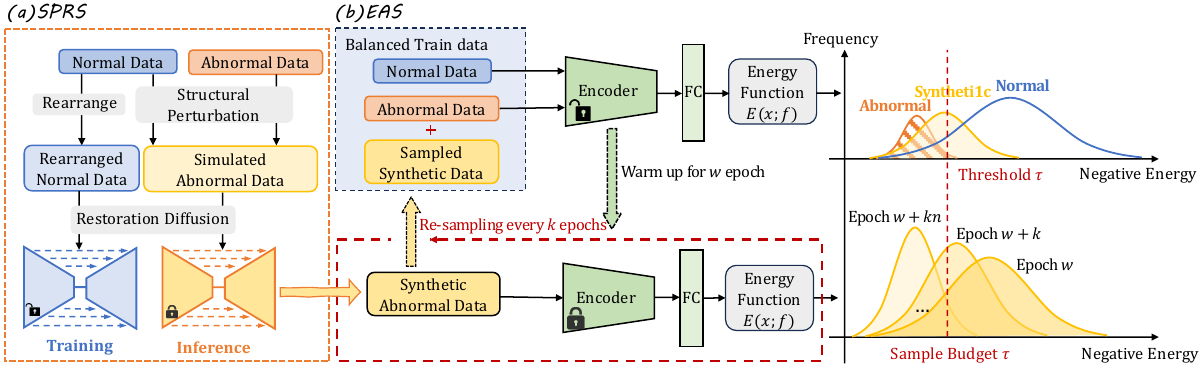}
    \caption{Framework of the proposed Perturb-and-Restore (P\&R) method. (a) The SPRS module generates synthetic abnormal chromosomes by simulating structural anomalies based on prior abnormality knowledge. (b) The EAS module performs dynamic online sampling of synthetic data during training. By referencing the energy distributions of real normal and abnormal samples, the model selects high-quality synthetic anomalies to enhance detection performance.}
    \label{fig:framework}
\end{figure*}

\subsection{Energy-guided Adaptive Sampling (EAS)}\label{sec:eas}
Energy-based learning models the data distribution by defining an energy function that assigns low energy values to probable data points and high energy values to improbable ones, with the energy score being theoretically aligned with the input's probability density. Inspired by this principle, we construct an energy-based anomaly detector to explicitly separate normal and abnormal chromosomes, and introduce a dynamic sampling strategy that leverages energy scores of real anomalies to iteratively select high-quality synthetic samples from the generated abnormal dataset $D_c^{ab}$.

\subsubsection{Energy-based Model}
For every chromosome category $c$, given its normal and abnormal sets $D_c^n$ and $D_c^{ab}$, we define an energy-based anomaly detector $f_\psi(x)$ that maps an input $x$ to a set of logits over binary classes $\mathcal{Y}=\{0,1\}$, where class $0$ denotes normal and class $1$ denotes abnormal. Without modifying $f_\psi(x)$, the free energy function $E(x;f)$ can be expressed in terms of the denominator of the softmax activation function as: 
\begin{equation} 
E(x;f) = -t \cdot \log \sum_{y=0}^{\mathcal{Y}}e^{f^y(x)/t}, 
\end{equation}
where $t$ is the temperature parameter, and $f^{y}(x)$ represents the logit for class $y$. This formulation assigns lower energy values to normal data and higher energy values to abnormal ones, an explicit energy gap is created, improving the discriminative ability of the anomaly detection model~\cite{liu2020energy}. To further reinforce this distinction, we define the margin-based energy loss function as:
\begin{equation} 
\begin{aligned}
\mathcal{L}_{\text{energy}}(\psi) &: = \mathbb{E}_{x_n \sim D^n_{\text{c}}} \left[ \max(0, E(x_n;f_\psi) - m_n) \right]^2 \\ &+ \mathbb{E}_{{x_{ab} \sim D^{ab}_{\text{c}}}} \left[ \max(0, m_{ab} - E(x_{ab};f_\psi)) \right]^2,
\end{aligned}
\end{equation}
where $m_{n}$ and $m_{ab}$ are margin hyperparameters that control the separation between normal and abnormal classes. 

\subsubsection{Dynamic Online Sampling Strategy}

\begin{algorithm}
\caption{Overall pipeline of Energy-guided Adaptive Sampling (EAS)}
\label{alg:eas}
Input: Dataset $D_c = \{D_c^n, D_c^{ab}\}$; synthetic anomaly set $\hat{D}_c^{ab}$; total epochs $T$; warm-up period $w$; sampling interval $k$; momentum coefficient $m$ \\
Output: Final anomaly detection model parameters $\psi_T$ \\
Initialize: $\psi_0$; selected synthetic set $\tilde{D}_c^{ab} \leftarrow \varnothing$
\begin{algorithmic}[1]
\For{$\text{epoch} = 1$ to $T$}
    \Statex // Data Selection Phase (only after warm-up)
    \If{$\text{epoch} \geq w$ and $(\text{epoch} - w) \bmod k = 0$}
        \State Compute energy scores for $D_c^n$ and $D_c^{ab}$
        \State Estimate threshold $\tau$ via Eq.~\eqref{eq:threshold}
        \State Sample $\tilde{D}_c^{\text{sample}}$ via Eq.~\eqref{eq:sample}
        \State Update $\tilde{D}_c^{ab} \gets \tilde{D}_c^{ab} \cup \tilde{D}_c^{\text{sample}}$
        \State Set $\mathit{do\_momentum} \leftarrow \text{True}$ // enable momentum update
    \Else
        \State Set $\mathit{do\_momentum} \leftarrow \text{False}$
    \EndIf

    \Statex // Model Training Phase
    \State Train $\psi_{\text{epoch}}$ on $D_c \cup \tilde{D}_c^{ab}$ using Eq.~\eqref{eq:objective}

    \Statex // Momentum Update (if applicable)
    \If{$\mathit{do\_momentum}$}
        \State $\psi_{\text{epoch}} \gets m \cdot \psi_{\text{epoch}-1} + (1 - m) \cdot \psi_{\text{epoch}}$
    \EndIf
\EndFor
\end{algorithmic}
\end{algorithm}

To effectively utilize the synthetic data $\hat{D}_c^{ab}$ generated via the diffusion restoration module, we propose a dynamic sampling strategy that adaptively selects informative samples during training. Since synthetic samples lack corresponding ground-truth images, their quality cannot be directly assessed. Instead, we use energy scores of real anomalies as a reference. Specifically, within the pool of generated abnormal samples, we iteratively select those with energy scores close to those of real abnormal data. By continuously exposing the model to newly sampled effective abnormal samples, the training process encourages the network to learn more distinctive embeddings, thereby improving its ability to generalize to unseen anomalies.

At scheduled intervals during training, the model computes energy scores for real normal and abnormal samples to update the energy threshold. Based on this, the set of selected synthetic anomalies $\tilde{D}_c^{sample}$ is given by:
\begin{equation} \label{eq:sample}
\tilde{D}_c^{sample} = \{ \hat{x}_{ab} \mid E(\hat{x}_{ab};f_\psi) > \tau, \hat{x}_{ab} \in \hat{D}_c^{ab} \},
\end{equation}
where $\tau$ is an adaptive threshold determined by matching a predefined recall level:
\begin{equation}\label{eq:threshold}
    \tau = \arg\min_{t} \left| \frac{\sum_{x_i \sim D_c} \mathds{1}(E(x_i;f_{\psi}) > t, y_i = 1)}{\sum_{x_i \sim D_c} \mathds{1}(y_i = 1)} - recall\_level \right|
\end{equation}
Here, $\mathds{1}(\cdot)$ denotes the indicator function, which returns $1$ if the condition holds and $0$ otherwise. In practice, due to the large volume of normal data, we sample $2,000$ normal chromosomes to compute the energy scores for reducing the computational cost and inference time. We summarize the full training pipeline of EAS in Algorithm~\ref{alg:eas}, which includes threshold estimation and iterative sampling integrated with model updates. 

This dynamic sampling strategy can be seamlessly integrated into the training process, ensuring that the model is continuously exposed to more informative synthetic anomalies, thereby improving anomaly detection performance. 

The overall training loss for EAS combines classification and energy objective:
\begin{equation} \label{eq:objective}
\mathcal{L} = \mathbb{E}_{(x,y) \sim D_c} \left[ -\log p(y|x) \right] + \lambda \cdot \mathcal{L}_{\text{energy}}
\end{equation}
where the first term represents the cross-entropy loss, $\lambda$ is a scaling hyperparameter that controls the contribution of the energy-based regularization.

\section{Experiments and results}

\subsection{Datasets}
\begin{table}[]
    \centering
    \caption{Summary of the chromosome used in this study. Each \textbf{category} corresponds to one of the 24 chromosome types (1–22, X, and Y). Values in the \textbf{train} column represent the number of abnormal versus normal chromosomes (\textit{Abnormal:Normal}). The \textbf{validation} and \textbf{test} sets each include 200 and 800 normal samples per category. The \textbf{unbalance ratio} quantifies the level of class imbalance (\textit{Normal/Abnormal}), while the \textbf{synthetic abnormality} column lists the number of synthetic abnormal chromosomes} 
    \label{tab:dataset_stats}
    \renewcommand{\arraystretch}{1}
    \rowcolors{3}{gray!15}{white}
    \setlength{\tabcolsep}{2pt}  
    \resizebox{0.5\textwidth}{!}{
    \begin{tabular}{c|c|c|c|c|c}
        \toprule
        \textbf{Category} & \textbf{Train} & \makecell{\textbf{Validation} \\ (200 Normal)} & \makecell{\textbf{Test} \\ (800 Normal)} & \makecell{\textbf{Unbalance}\\ \textbf{Ratio}} & \makecell{\textbf{Synthetic}\\ \textbf{Abnormality}}\\
        \midrule
        1  & 343:10227  & 57 & 170  & 29.81 & 7926 \\
        2  & 72:10217   & 12 & 33  & 141.90 & 8438\\
        3  & 42:10276   & 7  & 20  & 244.67 & 8799\\
        4  & 51:10275   & 8  & 25  & 201.47 & 9136\\
        5  & 44:10273  & 7  & 22  & 233.48 & 9269\\
        6  & 83:10271   & 14 & 40   & 123.75 & 8913\\
        7  & 70:10266   & 11  & 34   & 146.66 & 8870\\
        8  & 59:10276   & 10  & 29  & 174.17 & 9153\\
        9  & 468:10213  & 77  & 228 & 21.82  & 8741\\
        10 & 56:10281   & 9   & 27   & 189.59 & 9210\\
        11 & 46:10276   & 7   & 21  & 223.39 & 8947\\
        12 & 31:10277   & 6  & 16   & 331.52 & 9071\\
        13 & 124:10275  & 20  & 60   & 82.86  & 9700\\
        14 & 130:10396  & 21  & 63  & 79.97  & 9668\\
        15 & 100:10271   & 17  & 49   & 102.71 & 9593\\
        16 & 184:10281  & 30  & 90   & 55.88  & 9138\\
        17 & 31:10280   & 5   & 15   & 331.61 & 9179\\
        18 & 33:10267   & 6   & 17   & 311.12 & 9640\\
        19 & 22:10306   & 4  & 11   & 468.45 & 9561\\
        20 & 22:10281   & 4   & 10   & 467.32 & 9384\\
        21 & 101:10277   & 16  & 49   & 101.75 & 8600\\
        22 & 102:10280   & 17  & 50   & 100.78 & 9376\\
        X  & 49:7399    & 8   & 24  & 151.00 & 6244\\
        Y  & 304:2162   & 50  & 149  & 7.11  & 1835 \\
        \midrule
        \rowcolor{gray!30}
        \textbf{Total} & \textbf{2567:235657} & \textbf{423:4800} & \textbf{1252:19200} & - & \textbf{208391}\\
        \rowcolor{gray!30}
        \textbf{Ave.} & - & - & - & \textbf{179.87} & -\\
        \bottomrule
    \end{tabular}}
\end{table}

Building on our previous work~\cite{yang2025inclusive}, we collected an extensive dataset from Guangdong Provincial Maternal and Child Health Hospital, comprising over 260,000 chromosome images across all 24 categories. The dataset is split into training, validation, and test sets for anomaly detection. As shown in Table~\ref{tab:dataset_stats}, the training set exhibits a strong imbalance (4,242 abnormal vs. 259,657 normal samples, avg. ratio 179.87), mirroring real-world distributions.The validation and test sets are standardized, containing 200 normal samples for validation and 800 for testing, ensuring balanced evaluation. The table also reports the number of synthetic anomalies generated by our framework. For SPRS training, we created 212,392 pairs of original and rearranged chromosomes (from both normal and abnormal samples) without simulating anomalies. Low-quality rearrangements (MA score $<$ 85) were filtered out. The data was split into 98\% training (208,013 pairs) and 2\% test (4,379 pairs) for evaluating restoration performance.
\begin{table*}[]
\centering
\caption{ \centering Performance comparison of different methods for chromosomal anomaly detection (mean(std) \%). \\ \centering The best results and the second-best results are highlighted in \textbf{Bold} and in \underline{underline}.}
\label{tab:performance_comparison}
\renewcommand{\arraystretch}{1.1}
\setlength{\tabcolsep}{10pt}
\scriptsize
\resizebox{1.0\textwidth}{!}{
\begin{tabular}{lccccccc}
    \toprule
    \textbf{Methods} & \textbf{Acc}$\uparrow$ & \textbf{Sen}$\uparrow$ & \textbf{Spe}$\uparrow$ & \textbf{Pre(ab)}$\uparrow$ & \textbf{Pre(n)}$\uparrow$ & \textbf{AUC}$\uparrow$ & \textbf{F1}$\uparrow$ \\
    \midrule
    \multicolumn{8}{l}{\textbf{ResNet18}} \\ \hline
    \rowcolor{gray!15} Baseline & 95.94 (2.79) & 30.13 (31.21) & \textbf{99.22 (1.76)} & 65.99 (44.67) & 96.37 (2.23) & 79.36 (11.79) & 33.92 (31.20) \\ 
    Under-sampling & 91.39 (7.17) & 50.38 (22.34) & 93.39 (7.41) & 41.43 (26.68) & 97.22 (2.19) & 83.99 (9.46) & 38.27 (18.94) \\
    Over-sampling & 96.11 (3.28) & 53.63 (15.98) & 98.12 (2.59) & 71.17 (19.67) & 97.57 (1.75) & 91.44 (4.87) & 58.75 (13.32) \\
    Focal Loss & 95.33 (4.91) & 54.52 (16.50) & 97.96 (3.81) & 70.88 (14.92) & 96.01 (7.86) & 91.16 (5.06) & 59.44 (11.46) \\
    LDAM-DRW & 95.85 (4.70) & 54.97 (15.64) & 97.56 (5.04) & 68.63 (13.48) & \underline{97.77 (1.38)} & 91.18 (4.65) & \underline{59.59 (11.83)} \\
    BBN & 86.16 (16.11) & \underline{61.14 (30.01)} & 87.90 (17.69) & 43.11 (34.24) & 97.48 (1.97) & 86.74 (8.64) & 37.13 (24.66) \\
    BCL & 95.71 (3.34) & 57.04 (13.89) & 97.53 (3.15) & 60.24 (16.29) & 97.72 (1.78) & \underline{91.74} (4.50) & 57.34 (12.37) \\ \hline
    Energy-ood & 96.04 (3.57) & 54.58 (17.81) & 97.86 (3.59) & 68.69 (16.71) & 97.71 (1.49) & 91.72 (5.54) & 59.06 (14.25) \\  \hline
    ReAD & \underline{96.29 (3.09)} & 48.61 (18.75) & 98.41 (2.46) & \underline{73.86 (15.71)} & 97.49 (1.52) & 91.21 (4.37) & 56.17 (15.21) \\
    ChroSiameseAD & 96.04 (2.84) & 49.86 (18.61) & 98.27 (1.83) & 68.01 (17.64) & 97.37 (2.05) & 87.40 (5.97) & 54.93 (13.90) \\
    \rowcolor{gray!30} P\&R (Ours) & \textbf{97.23 (2.98)} \inc $_{\textcolor{red}{0.94}}$& \textbf{70.06 (12.02)} \inc $_{\textcolor{red}{8.92}}$& \underline{98.44 (2.79)} \dec $_{\textcolor{darkgreen}{0.78}}$& \textbf{78.24 (11.97)} \inc $_{\textcolor{red}{4.38}}$& \textbf{98.48 (1.07)}  \inc $_{\textcolor{red}{0.71}}$& \textbf{95.00 (2.75)} \inc $_{\textcolor{red}{3.26}}$ & \textbf{73.03 (9.54)} \inc $_{\textcolor{red}{13.44}}$\\
    \midrule
    \multicolumn{8}{l}{\textbf{ResNet50}} \\ \hline
    \rowcolor{gray!15} Baseline & 95.94 (2.79) & 29.28 (31.21) & \textbf{99.36 (1.76)} & 53.09 (44.67) & 96.30 (2.23) & 81.26 (11.79) & 30.89 (31.20) \\ 
    Under-sampling & 93.31 (4.81) & 40.58 (23.61) & 95.73 (5.02) & 42.17 (25.06) & 96.94 (2.04) & 82.07 (11.15) & 36.78 (18.78) \\
    Over-sampling & 95.58 (4.02) & 54.55 (19.08) & 97.46 (3.95) & 59.52 (15.77) & 97.68 (1.60) & 90.49 (5.05) & 55.30 (15.99) \\
    Focal Loss & 95.36 (4.57) & 55.57 (16.60) & 97.20 (4.83) & 62.03 (21.10) & 97.64 (1.72) &  \underline{91.41 (4.62)} & 56.00 (14.68) \\
    LDAM-DRW & 95.64 (4.23) & 57.25 (18.19) & 97.29 (3.91) & 65.01 (15.22) &  \underline{97.78 (1.63)} & 91.31 (4.45) & \underline{58.67 (13.71)} \\
    BBN & 73.42 (22.03) & 60.76 (35.55) & 74.87 (24.75) & 38.21 (41.80) & 97.21 (3.62) & 82.55 (7.75) & 21.78 (19.10) \\
    BCL & 94.80 (3.67) & \underline{61.24 (11.03)} & 96.51 (3.08) & 51.45 (15.95) & 97.76 (2.02) & 90.51 (4.75) & 54.74 (11.94) \\ \hline 
    Energy-ood & 95.83 (3.73) & 50.75 (18.70) & 97.83 (3.86) & 67.47 (23.69) & 96.38 (6.20) & 91.01 (4.76) & 55.53 (16.49) \\ \hline
    ReAD & \underline{96.21 (2.55)} & 51.07 (18.10) & 98.38 (1.98) & \underline{70.67 (19.32)} & 97.49 (1.64) & 90.23 (5.03) & 56.55 (15.97) \\
    ChroSiameseAD & 95.82 (3.28) & 46.12 (18.59) & 98.31 (2.82) & 69.46 (20.95) & 97.19 (1.89) & 83.73 (9.87) & 52.60 (16.34) \\ 
    \rowcolor{gray!30} P\&R (Ours) & \textbf{97.19 (3.49)} \inc $_{\textcolor{red}{0.98}}$ & \textbf{68.35 (11.61)} \inc $_{\textcolor{red}{7.11}}$& \underline{98.40 (3.53)} \dec $_{\textcolor{darkgreen}{0.96}}$& \textbf{79.56 (13.69)} \inc $_{\textcolor{red}{8.89}}$ & \textbf{98.37 (1.04)} \inc $_{\textcolor{red}{0.59}}$& \textbf{94.83 (3.19)} \inc $_{\textcolor{red}{3.42}}$& \textbf{72.46 (9.17)} \inc $_{\textcolor{red}{13.79}}$\\
    \bottomrule
\end{tabular}}
\end{table*}
\subsection{Implementation Details and Evaluation Metrics}
\subsubsection{Implementation}

Our algorithm is implemented in Python with PyTorch and runs on an NVIDIA V100 GPU. In \textbf{SPRS}, chromosome segments are rearranged with interval $l$ randomly sampled from $\frac{1}{20}$ to $\frac{1}{5}$ of the chromosome height. The MAC score is computed with the sample number $m = 6$. For restoration, we adopt the mean-reverting SDE from~\cite{luo2023image} with a cosine noise schedule ($\sigma = 10$, 100 diffusion steps). Images are resized to $256 \times 256$. We train using Adam (learning rate $1\times10^{-4}$, $\beta_1{=}0.9$, $\beta_2{=}0.99$, batch size 8) for 300K iterations, decaying the learning rate to $1\times10^{-5}$ at 200K. In \textbf{EAS}, ResNet18 and ResNet50~\cite{he2016deep} are used as backbones. The temperature is $t = 1$, and the margins are set to $m_n = -27$, $m_{ab} = -5$. The model is trained for 100 epochs with a 10-epoch warm-up and sampling interval $k = 10$. Cosine annealing is used for learning rate decay from an initial value of $5\times10^{-3}$, with a batch size of 64. The recall threshold in Eq.~\eqref{eq:threshold} is 0.7, and the loss weight $\lambda$ in Eq.~\eqref{eq:objective} is set to 0.1.

\subsubsection{Evaluation Metrics}.
For anomaly detection, we report accuracy (Acc), sensitivity (Sen), specificity (Spe), precision (Pre), F1-score (F1), and AUC. F1-score is especially useful under class imbalance, as it balances precision and recall. In SPRS, we evaluate the restoration performance using peak signal-to-noise ratio (PSNR), structural similarity index measure (SSIM) to measure pixel-level and perceptual similarity, and kernel inception distance (KID) to assess distributional divergence from real data. Higher PSNR/SSIM and lower KID indicate better visual quality and realism. 

\subsection{Comparisons with State-of-the-arts}

We compare our method against three categories of approaches using ResNet18 and ResNet50 backbones: (1) \textbf{Long-tailed learning methods}, including sampling strategies (over-sampling~\cite{ando2017deep}, under-sampling~\cite{he2009learning}), re-weighting losses (focal loss~\cite{ross2017focal}, LDAM-DRW~\cite{cao2019learning}), and architecture enhancements (BBN~\cite{zhou2020bbn}, BCL~\cite{zhu2022balanced}); (2) \textbf{Energy-based methods}, represented by energy-ood~\cite{liu2020energy}, which use energy scores to distinguish normal and abnormal samples; and (3) \textbf{Chromosomal anomaly detection methods}, including ReAD~\cite{cox2022automated} and ChroSiameseAD~\cite{bechar2023highly}.

As shown in Table~\ref{tab:performance_comparison}, baseline models exhibit low sensitivity, reflecting the difficulty of detecting rare abnormalities under imbalance. Our P\&R method achieves the best overall performance on both backbones, with substantial gains in sensitivity (+8.92 for ResNet18, +7.11 for ResNet50), abnormal precision (+4.38, +8.89), and F1-score (+13.44, +13.79). Although specificity slightly decreases, the drop is minimal, indicating an improved trade-off between anomaly detection (higher recall) and false positives. Notably, ReAD and ChroSiameseAD perform worse than general long-tailed methods, likely due to their limited scope and lack of imbalance handling, which hampers generalization across diverse chromosomal anomalies.

\begin{figure*}
    \centering
    \includegraphics[width=1\linewidth]{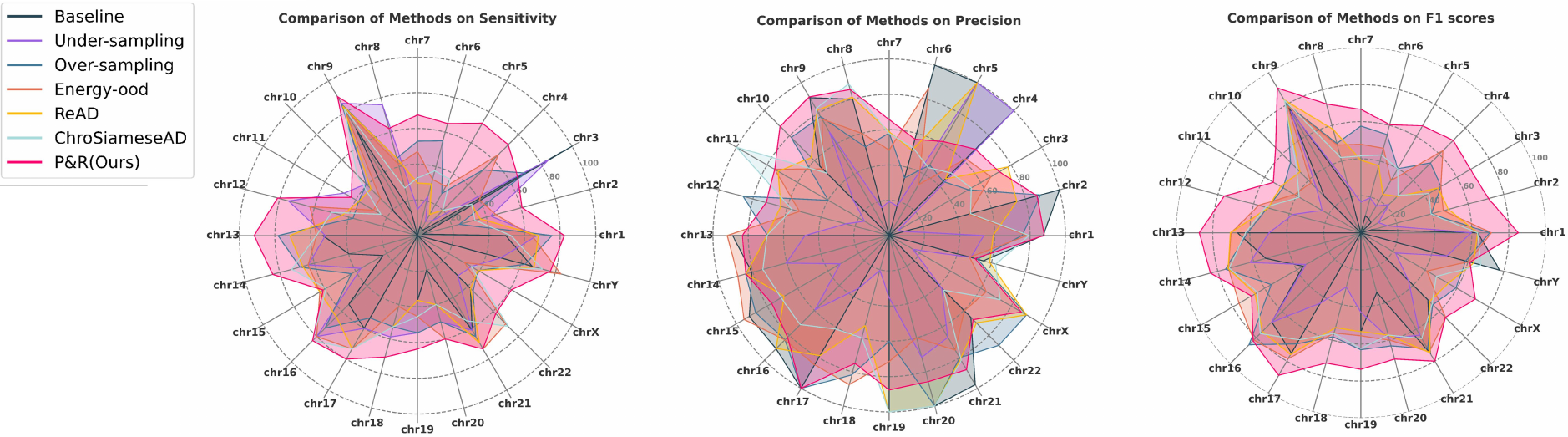}
    \caption{Radar chart of chromosomal anomaly detection on 24 categories (sensitivity, precision of abnormality, and F1 score).}
    \label{fig:radar_chart}
\end{figure*}

To further demonstrate the effectiveness of our method, Fig.~\ref{fig:radar_chart} presents a radar chart of sensitivity, abnormal precision, and F1-score across 24 chromosome categories. P\&R consistently outperforms other methods in most categories. Under extreme imbalance, baseline models often show degraded sensitivity and abnormal precision dropping to 0 or rising to 100—indicating either complete failure to detect anomalies or severe overfitting to a few patterns. By introducing diverse synthetic anomalies, P\&R not only balances the dataset but also enriches abnormal pattern variability, enabling more robust learning and substantially improving detection across nearly all categories.

\begin{figure*}
    \centering    \includegraphics[width=1\linewidth]{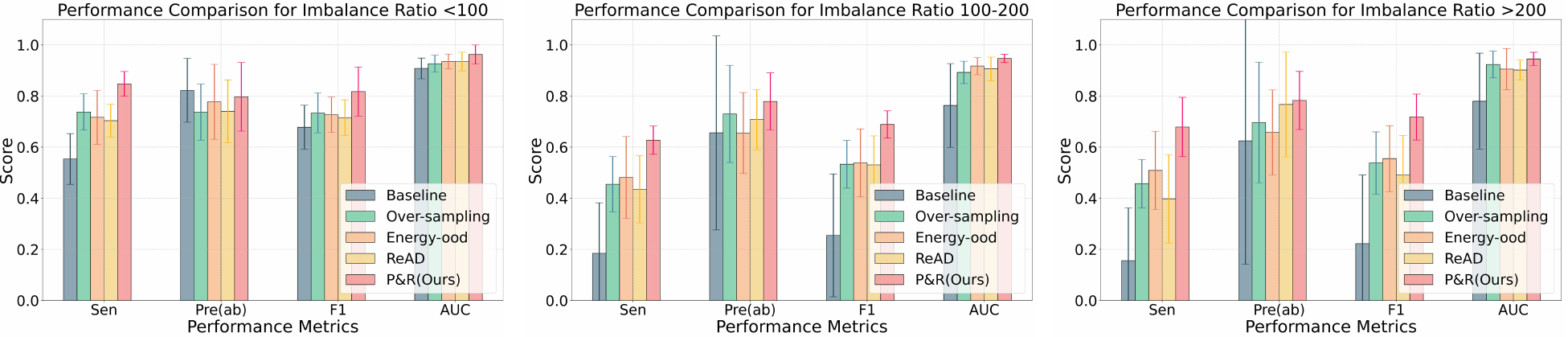}
    \caption{Performance comparisons under different imbalance ratio levels.}
    \label{fig:imbalance_ratio}
\end{figure*}

As shown in Fig.~\ref{fig:imbalance_ratio}, we group the 24 chromosome categories into three subgroups based on imbalance ratios. When the ratio exceeds 200, the baseline exhibits large error bars, indicating that data scarcity severely restricts the network’s ability to learn meaningful patterns. As the imbalance increases, the performance gains of other methods also become more evident. Notably, P\&R demonstrates even greater improvements, particularly in sensitivity for the 100-200 and $>$ 200 subgroups. This highlights P\&R's effectiveness in detecting anomalies, especially in scenarios with extreme data imbalance.

\subsection{Ablation Study}
\subsubsection{Component validation}

In Table \ref{tab:ablation}, we ablate the designs introduced in Section \ref{sec:sprs} and Section \ref{sec:eas}, which are proposed to generate abnormal chromosomes (before and after restoration, SYN and SYN*) and sampling high-quality synthetic samples (EAS). The results indicate that using synthetic data without restoration, as described in Section \ref{sec:structural_perturbation} (SYN), yields only a marginal improvement. In contrast, incorporating rearrangement restoration (SYN⁎) significantly enhances model performance, particularly in sensitivity, precision, F1 and AUC, across both ResNet18 and ResNet50 architectures. Statistical significance, as indicated by $p_1$ (relative to the baseline), confirms the effectiveness of the synthetic data after restoration. Furthermore, the addition of EAS further improves classification results, particularly when using SYN data ($p_2 < 0.05$), suggesting that when the quality of synthetic data is suboptimal, EAS effectively samples data that closely resemble real abnormal chromosomes, thereby enhancing model performance. Overall, the combination of synthetic data restoration and EAS leads to superior classification accuracy, validating the proposed methodologies.

\begin{table*}[]
\centering
\caption{Performance comparison with different components. SYN represents the abnormality simulation data before restoration, while SYN* denotes the synthetic data after restoration. $p_1$ indicates the p-value of the paired t-test relative to the baseline, while $p_2$ represents the p-value relative to the model without EAS.}
\label{tab:ablation}
\renewcommand{\arraystretch}{1.1}
\setlength{\tabcolsep}{10pt}
\resizebox{1.0\textwidth}{!}{
\begin{tabular}{ccccccccccccc}
    \toprule
     \textbf{SYN} & \textbf{SYN*} & \textbf{EAS} & \textbf{Acc}$\uparrow$ & \textbf{Sen}$\uparrow$ & \textbf{Spe}$\uparrow$ & \textbf{Pre(ab)}$\uparrow$ & \textbf{Pre(n)}$\uparrow$ & \textbf{AUC}$\uparrow$ & \textbf{F1}$\uparrow$ & \textbf{$p_1$} & \textbf{$p_2$} \\
    \midrule
    \multicolumn{12}{l}{\textbf{ResNet18}} \\ \hline
    \rowcolor{gray!15} & & & 95.94 (2.79) & 30.13 (31.21) & \textbf{99.22 (1.76)} & 65.99 (44.67) & 96.37 (2.23) & 79.36 (11.79) & 33.92 (31.20) & $^{\star}$ & - \\
    \checkmark & & & 95.91 (3.00) & 30.65 (24.13) & 95.34 (16.88) & 63.75 (28.18) & 96.81 (1.57) & 84.42 (10.38) & 37.77 (24.79) & $^{\star}$ 0.328 & $^{\vartriangle}$ \\
    \checkmark & & \checkmark & 96.42 (3.19) & 55.30 (16.25) & 98.42 (2.49) & 71.90 (12.30) & 97.65 (1.97) & 90.75 (4.37) & 60.88 (12.05) & $^{\star}$ $4.24\times 10^{-5}$ & $^{\vartriangle}$ $3.77\times 10^{-6}$ \\
    & \checkmark & & \textbf{97.38 (2.64)} & \underline{69.14 (16.90)} & \underline{98.58} (2.84) & \underline{74.85 (12.76)} & \textbf{98.54 (0.94)} & \textbf{95.94 (3.51)} & \underline{70.91 (13.62)} &$^{\star}$ $1.19\times 10^{-6}$  & $^{\triangledown}$\\
    \rowcolor{gray!30} & \checkmark & \checkmark & \underline{97.23 (2.98)} & \textbf{70.06 (12.04)} & 98.44 (2.79) & \textbf{78.24 (11.97)} & \underline{98.48 (1.07)} & \underline{95.00 (2.75)} & \textbf{73.03 (9.54)} & $^{\star}$ $1.51\times 10^{-7}$ & $^{\triangledown}$ 0.226 \\
    \midrule
    \multicolumn{12}{l}{\textbf{ResNet50}} \\ \hline
    \rowcolor{gray!15} & & & 95.94 (2.79) & 29.28 (31.21) & \textbf{99.36 (1.76)} & 53.09 (44.67) & 96.30 (2.23) & 81.26 (11.79) & 30.89 (31.20) & $^{\star}$ & - \\
    \checkmark & & & 96.54 (2.71) & 30.76 (37.20) & 99.13 (2.81) & 52.04 (41.85) & 96.75 (2.49) & 74.93 (19.79) & 29.46 (36.30) &$^{\star}$ 0.822 & $^{\vartriangle}$\\
    \checkmark & & \checkmark & 95.88 (3.85) & 59.32 (19.25) & 96.41 (7.09) & 65.84 (17.21) & 97.94 (1.50) & 88.11 (6.16) & 60.87 (14.27) & $^{\star}$ $4.71\times 10^{-6}$ & $^{\vartriangle}$ $5.29\times 10^{-5}$ \\
    & \checkmark & & \textbf{97.48 (2.01)} & \textbf{70.46 (13.60)} & \underline{98.65 (2.38)} & \underline{74.93 (13.42)} & \textbf{98.61 (0.79)} & \textbf{95.95 (3.05)} & \underline{71.69 (11.23)} & $^{\star}$ $2.29\times 10^{-7}$ & $^{\triangledown}$\\
    \rowcolor{gray!30} & \checkmark & \checkmark & \underline{97.19 (3.49)} & \underline{68.35 (11.61)} & 98.40 (3.53) & \textbf{79.56 (13.69)} & \underline{98.37 (1.04)} & \underline{94.83 (3.19)} & \textbf{72.46 (9.17)} & $^{\star}$ $1.52\times 10^{-7}$  & $^{\triangledown}$ 0.375 \\
    \bottomrule
\end{tabular}}
\end{table*}

\subsubsection{Generality of Synthetic Images}

In Table~\ref{tab:generality}, we evaluate the generalization capability of the synthetic data (SYN⁎) generated by the APRS module across multiple baseline methods. The results demonstrate that incorporating SYN⁎ data into the training process across various methods, including standard cross-entropy loss (CE), ReAD, and Energy-OOD, leads to a significant performance improvement, particularly in Sensitivity, Precision, and F1-score. Notably, CE benefits the most from SYN⁎ augmentation, with Sensitivity increasing by 39.01\% and F1-score improving by 36.99\%, indicating that the synthetic data effectively alleviates the data imbalance problem. ReAD and Energy-OOD also show substantial gains. Moreover, the consistent enhancement observed across different methods highlights the robustness and broad applicability of the synthetic data in improving model performance.

\begin{table}[h!]
    \centering
    \caption{Performance improvement across different methods with and without SYN*
}
    \label{tab:generality}
    \resizebox{0.5\textwidth}{!}{%
    \begin{tabular}{lcccccc}
        \toprule
        \textbf{Methods} & \textbf{Acc}$\uparrow$ & \textbf{Pre(ab)}$\uparrow$ & \textbf{Sen}$\uparrow$ & \textbf{F1}$\uparrow$ & \textbf{AUC}$\uparrow$ \\
        \midrule
        CE             & 95.94 (2.79) & 65.99 (44.67) & 30.13 (31.21) & 33.92 (31.20) & 79.36 (11.79) \\
        CE+SYN*        & 97.38 (2.64) \inc $_{\textcolor{red}{1.44}}$ & 74.85 (12.76) \inc $_{\textcolor{red}{8.86}}$ & 69.14 (16.90) \inc $_{\textcolor{red}{39.01}}$& 70.91 (13.62) \inc $_{\textcolor{red}{36.99}}$ & 95.94 (3.51) \inc $_{\textcolor{red}{16.58}}$\\
        \midrule
        ReAD           & 96.29 (3.09) & 73.86 (15.71) & 48.61 (18.75) & 56.17 (15.21) & 91.21 (4.37) \\
        ReAD+SYN*      & 96.58 (2.67) \inc $_{\textcolor{red}{0.29}}$ & 76.87 (15.31) \inc $_{\textcolor{red}{3.01}}$& 53.41 (18.11) \inc $_{\textcolor{red}{4.8}}$& 60.67 (14.60) \inc $_{\textcolor{red}{4.5}}$& 91.39 (4.79) \inc $_{\textcolor{red}{0.18}}$\\
        \midrule
        Energy\_ood    & 96.04 (3.57) & 68.69 (16.71) & 54.58 (17.81) & 58.94 (14.39) & 91.72 (5.54) \\
        Energy\_ood+SYN* & 97.47 (2.60) \inc $_{\textcolor{red}{1.43}}$& 73.77 (13.05) \inc $_{\textcolor{red}{5.08}}$& 71.09 (17.44) \inc $_{\textcolor{red}{16.51}}$& 71.64 (14.59) \inc $_{\textcolor{red}{12.7}}$& 95.12 (3.78) \inc $_{\textcolor{red}{3.4}}$\\
        \bottomrule
    \end{tabular}
    }
\end{table}

\subsubsection{Settings of Hyperparameters in EAS}

Table~\ref{tab:hyper} examines the impact of loss weighting coefficient $\lambda$ and sampling interval $k$. For $\lambda$, we observe a trade-off between \textbf{precision} and \textbf{sensitivity}: increasing $ \lambda $ boosts precision (highest at 83.48 for $ \lambda=10 $) but lowers sensitivity (65.12), while smaller values like $ \lambda = 0.1 $ offer a more balanced performance across all metrics, achieving the best F1-score (73.03). Regarding \( k \), both \( k=10 \) and \( k=15 \) yield strong performance. Specifically, \( k=10 \) favors precision (78.24), whereas \( k=15 \) achieves the highest sensitivity (74.33), F1-score (74.01), and AUC (95.54). Shorter intervals (e.g., \( k=5 \)) slightly increase sensitivity but reduce precision, while overly large intervals (\( k=20 \)) degrade overall performance. Overall, the results indicate that both $k = 10$ and $k = 15$ are effective choices, with $k = 10$ favoring precision and $k = 15$ favoring sensitivity, providing a flexible trade-off for optimizing the model's anomaly detection performance.

\begin{table}[h!]
    \centering
    \caption{Anomaly detection performance under varying loss weights ($\lambda$) and sampling intervals ($k$).}
    \label{tab:hyper}
    \resizebox{0.5\textwidth}{!}{%
    \begin{tabular}{lccccc}
        \toprule
        & \textbf{Acc}$\uparrow$ & \textbf{Pre(ab)}$\uparrow$ & \textbf{Sen}$\uparrow$ & \textbf{F1}$\uparrow$ & \textbf{AUC}$\uparrow$ \\
        \midrule
        \multicolumn{6}{l}{\textbf{Coefficient $\lambda$}} \\
        0.1 & 97.23 (2.98) & 78.24 (11.97) & \textbf{70.06 (12.04)} & \textbf{73.03 (9.54)} & \textbf{95.00 (2.75)} \\
        1   & 97.26 (2.22) & 74.67 (10.97) & 69.87 (10.53) & 71.89 (9.79) & 94.09 (3.48) \\
        10  & \textbf{97.61 (2.46)} & \textbf{83.48 (10.48)} & 65.12 (15.07) & 72.30 (12.12) & 92.08 (5.77) \\
        \midrule
        \multicolumn{6}{l}{\textbf{Interval $k$}} \\
        5   & 96.74 (4.37) & 70.55 (13.28) & 73.84 (13.28) & 71.54 (10.68) & 94.77 (3.56) \\
        10  & \textbf{97.23 (2.98)} & \textbf{78.24 (11.97)} & 70.06 (12.04) & 73.03 (9.54) & 95.00 (2.75) \\
        15  & 97.16 (3.40) & 75.70 (10.78) & \textbf{74.33 (11.51)} & \textbf{74.01 (8.11)} & \textbf{95.54 (3.61)} \\
        20  & 97.09 (3.28) & 72.10 (13.28) & 74.14 (11.95) & 72.32 (10.40) & 95.46 (3.30) \\
        \bottomrule
    \end{tabular}
    }
\end{table}

\subsection{Analysis of Label-Free Restoration at Different Iterations}

Due to the lack of ground-truth labels for synthetic abnormalities, we evaluate the restoration model using: (1) image fidelity (PSNR, SSIM), computed on the test set during restoration training, assesses the model's ability to recover visual details from rearranged inputs; (2) distribution alignment (KID), measures how closely the synthetic images resemble real normal and abnormal data distributions; (3) downstream detection performance, reflects the practical utility of the synthetic data in improving anomaly classification.

As shown in Table~\ref{tab:restoration}, PSNR and SSIM steadily improve across training iterations, indicating better recovery of visual details. However, this primarily reflects the reconstruction of normal structures, as the test set is dominated by normal chromosomes. Anomaly detection accuracy peaks at 50K iterations, then declines—suggesting that the restoration model may overfit to normal structures, reducing the diversity or realism of the synthesized anomalies. Furthermore, KID values decrease for both normal and abnormal references as training progresses. As shown in the Table, the reduction is more pronounced for normal data, likely because most training pairs originate from normal chromosomes—biasing the restoration model toward recovering normal structures. Notably, the KID scores relative to normal data at 10K (0.0324) and 50K (0.0325) iterations are closest to that at iteration 0 (0.0412, the divergence between real normal and abnormal chromosomes), suggesting that the synthetic anomalies at these stages better approximate the true abnormal–normal separation. In particular, 50K achieves a lower KID (0.0342) relative to abnormal data than 10K (0.0460), indicating higher realism and contributing to its superior anomaly detection results.

\begin{table}[]
    \centering
    \caption{Restoration and detection performance at different iterations of training restoration model. For KID, the value at Iter~0 represents the divergence between real abnormal and normal chromosome images before restoration training. }
    \label{tab:restoration}
    \resizebox{0.4\textwidth}{!}{%
    \begin{tabular}{ccccc}
        \toprule
         & \multicolumn{2}{c}{\textbf{Restoration}} & \multicolumn{2}{c}{\textbf{KID}$\downarrow$} \\
        \cmidrule(lr){2-3} \cmidrule(lr){4-5}
        \textbf{Iter} & \textbf{PSNR$\uparrow$} & \textbf{SSIM$\uparrow$} & \textbf{vs. Normal} & \textbf{vs. Abnormal} \\
        \midrule
        0      &  -     &  -      & 0.0412 & 0 \\
        10K    & 29.94 & 0.9687 & 0.0324 & 0.0460 \\
        50K    & 30.51 & 0.9678 & 0.0305 & 0.0342 \\
        100K   & 31.66 & 0.9708 & 0.0159 & 0.0318 \\
        150K   & 33.68 & 0.9768 & 0.0144 & 0.0325 \\
        200K   & 34.92 & 0.9805 & 0.0132 & 0.0307 \\
        \midrule
        & \multicolumn{4}{c}{\textbf{Anomaly Detection}} \\
        \cmidrule(lr){2-5}
        \textbf{Iter} & \textbf{Acc$\uparrow$} & \textbf{Sen$\uparrow$} & \textbf{F1$\uparrow$} & \textbf{AUC$\uparrow$} \\
        \midrule
        10K    & \textbf{97.34} & 63.10 & 70.91 & 93.91 \\
        50K    & 97.23 & 70.06 & \textbf{73.03} & \textbf{95.00} \\
        100K   & 96.49 & 70.99 & 65.17 & 93.85 \\
        150K   & 96.31 & 69.06 & 65.23 & 92.45 \\
        200K   & 96.53 & \textbf{72.42} & 67.04 & 93.88 \\
        \bottomrule
    \end{tabular}
    }
\end{table}

\subsection{User Study on Synthetic Image Realism}

To assess the perceptual realism of synthetic images, we conducted a user study with 14 participants: 7 non-experts (Group 1) and 7 cytogenetics experts (Group 2). Each participant viewed a randomized set of 100 chromosome images (50 real, 50 synthetic) and labeled each as real or synthetic based on visual inspection. As shown in Table~\ref{tab:userstudy}, experts showed only marginally higher accuracy than non-experts, with no significant difference ($p = 0.779$), suggesting domain knowledge had limited impact. When responses were aggregated per image, users identified real images more accurately (62.6\%) than synthetic ones (43.0\%), a statistically significant difference ($p = 4.86 \times 10^{-9}$). This asymmetry suggests that synthetic images were often mistaken as real, demonstrating the visual realism of our generative model.

\begin{table}[h!]
    \centering
    \caption{Results of user study}
    \label{tab:userstudy}
    \resizebox{0.5\textwidth}{!}{%
    \begin{tabular}{lccc}
        \toprule
        \textbf{Acc} $\uparrow$ & \textbf{Group 1} & \textbf{Group 2} & \textbf{$p$-value} \\
        \midrule
        Per-user Acc & 52.29 (5.20) & 53.29 (6.76) & 0.779 \\ 
        \midrule
        \textbf{Acc} $\uparrow$ & \textbf{Real Images} & \textbf{Synthetic Images} & \textbf{$p$-value} \\
        \midrule
        Per-image Acc & 62.57 (15.36) & 43.00 (14.81) & $4.86\times 10^{-9}$ \\
        \bottomrule
    \end{tabular}
    }
\end{table}

\subsection{Analysis of Energy-guided Adaptive Sampling}

\begin{figure*}
    \centering
\includegraphics[width=1.0\linewidth]{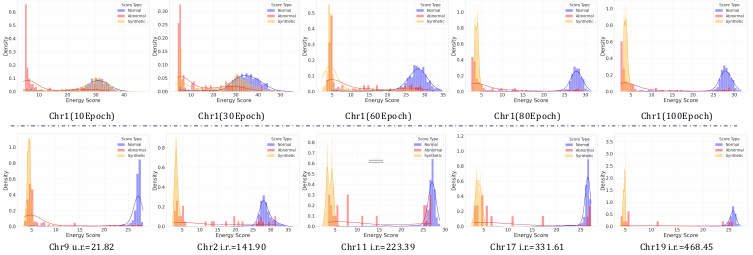}
    \caption{Visualization of Energy score distributions for normal, abnormal, and synthetic data. (Top) Evolution of energy distributions for Chr1 over training epochs, showing progressive separation between normal (blue) and abnormal (red and orange) samples. (Bottom) Final energy score distributions across representative chromosomes with varying imbalance ratios (i.r.), demonstrating that the proposed EAS module maintains clear separability between normal and abnormal classes even under extreme imbalance conditions (please zoom in to view finer details).}
    \label{fig:energy_score}
\end{figure*}

To analyze the effect of the online sampling strategy, we visualize the energy score distribution for normal, abnormal, and synthetic data, respectively. From the first row in Fig. \ref{fig:energy_score}, we observe that at the beginning of training, the network struggles to distinguish between normal and abnormal data, with the energy scores of synthetic data widely distributed around both normal and abnormal samples, especially clustering near normal samples. This indicates that the network initially has weak anomaly recognition capability, and a substantial portion of synthetic data may confuse the model. As training progresses and more synthetic samples with energy scores closer to the abnormal distribution are selected, the availability of useful synthetic data increases, further enhancing the network's classification ability. By the later stages of training, the distributions of normal and abnormal energy scores become more clearly separated, demonstrating improved discrimination. The second row of Fig. \ref{fig:energy_score} presents the final epoch’s distribution under different imbalance ratios, showing that our method maintains strong separability between normal and abnormal classes across various levels of data imbalance.

\section{Conclusion and future work}
In this work, we proposed \textbf{Perturb-and-Restore (P\&R)}, a novel simulation-driven structural augmentation framework for addressing the long-tailed distribution in structural chromosomal anomaly detection. The \textbf{Structure Perturbation and Restoration Simulation (SPRS)} module generates diverse abnormal patterns via perturbation of normal chromosomes and refines them through a diffusion-based restoration model, eliminating the reliance on real abnormal samples. The \textbf{Energy-guided Adaptive Sampling (EAS)} module further improves performance by dynamically selecting high-quality synthetic anomalies based on energy scores. Extensive experiments across 24 chromosome categories demonstrate that P\&R achieves state-of-the-art performance under severe data imbalance. While P\&R demonstrates strong performance, there remains ample room for exploration. Incorporating biological priors or clinically informed mutation patterns could further improve simulation fidelity. We hope this work encourages broader research into AI-assisted karyotyping and data-efficient anomaly detection.

\section*{Acknowledgments}
This work was supported by the King Abdullah University of Science and Technology (KAUST) Office of Research Administration (ORA) under Award Nos. REI/1/5234-01-01, REI/1/5414-01-01, REI/1/5289-01-01, REI/1/5404-01-01, REI/1/5992-01-01, and URF/1/4663-01-01. Additional support was provided by the Center of Excellence for Smart Health (KCSH) under Award No. 5932, and the Center of Excellence on Generative AI under Award No. 5940.

\section*{Ethics Statement}
The anonymized partial data used in this study were reviewed and approved by the Ethics Committee of Guangdong Provincial Maternal and Child Health Hospital (Approval Number: 202301210). 

\bibliographystyle{IEEEtran}
\bibliography{mybibfile}

\end{document}